\documentclass[10pt, a4paper]{article}

\usepackage[final]{lrec-coling2024} 
\usepackage[title]{appendix}
\usepackage{graphicx}
\usepackage{subfig}
\usepackage{adjustbox}
\usepackage{multirow}
\usepackage{hyperref}

\title{SoftMCL: Soft Momentum Contrastive Learning for Fine-grained Sentiment-aware Pre-training}

\name{Jin Wang$^{\dag}$, Liang-Chih Yu$^{\ddag,}$* \thanks{*Corresponding author} and Xuejie Zhang$^{\dag}$}

\address{$^{\dag}$School of Information Science and Engineering, Yunnan University, Kunming, China \\
         $^{\ddag}$Department of Information Management, Yuan Ze University, Taoyuan, Taiwan \\
         \{wangjin, xjzhang\}@ynu.edu.cn, lcyu@saturn.yzu.edu.tw\\}

\abstract{
The pre-training for language models captures general language understanding but fails to distinguish the affective impact of a particular context to a specific word. Recent works have sought to introduce contrastive learning (CL) for sentiment-aware pre-training in acquiring affective information. Nevertheless, these methods present two significant limitations. First, the compatibility of the GPU memory often limits the number of negative samples, hindering the opportunities to learn good representations. In addition, using only a few sentiment polarities as hard labels, e.g., positive, neutral, and negative, to supervise CL will force all representations to converge to a few points, leading to the issue of latent space collapse. This study proposes a soft momentum contrastive learning (SoftMCL) for fine-grained sentiment-aware pre-training. Instead of hard labels, we introduce valence ratings as soft-label supervision for CL to fine-grained measure the sentiment similarities between samples. The proposed SoftMCL is conducted on both the word- and sentence-level to enhance the model's ability to learn affective information. A momentum queue was introduced to expand the contrastive samples, allowing storing and involving more negatives to overcome the limitations of hardware platforms. Extensive experiments were conducted on four different sentiment-related tasks, which demonstrates the effectiveness of the proposed SoftMCL method. The code and data of the proposed SoftMCL is available at: \href{https://www.github.com/wangjin0818/SoftMCL/}{https://www.github.com/wangjin0818/SoftMCL/}.
 \\ \newline  }

\begin{document}

\maketitleabstract

\section{Introduction}

Pre-trained language models (PLMs), such as BERT \cite{Devlin2018}, XLNet \cite{Yang2019a}, RoBERTa \cite{Liu2019a}, and DeBERTa \cite{He2021}, have proven their powerful ability to learn representations for natural language understanding and generation tasks. To learn language usage, these models were first trained on large-scale unlabeled corpora by unsupervised tasks, such as masked language model (MLM) and next sentence prediction (NSP). Then, they can be transferred to another task by supervised training on a smaller task-specific dataset. The great success of those PLMs is attributed to the practical pre-training tasks.

The core idea of pre-training is to generate similar contextual representations for words or sentences with similar contexts. Although the training paradigm works well for semantic-oriented applications with supervision for the role of language usage, they fail to distinguish the affective impact of a particular context to a specific word since certain words may have special semantics and sentiments in specific contexts \cite{Agrawal2018}. For example, the emotion of \underline{\emph{The battery life is long}} is positive, but \underline{\emph{It takes a long time to focus}} is negative, even though both sentences contain the word \underline{\emph{long}}. Similarly, \underline{\emph{The scope of his book is ambitious}}, and \underline{\emph{The government's decisions to begin the ambitious}} \underline{\emph{plans which cost a lot}}, contain the same \underline{\emph{ambitious}} vocabulary but clearly express the opposite attitude. As a result, PLMs are still sub-optimal in sentiment-related tasks due to the ignorance of the affective information in pre-training. As reported by \citet{Kassner2020b}, It is still hard for PLMs to handle words with contradiction sentiment or negation expression, which are critical in sentiment analysis tasks.

To assist PLMs in obtaining affective information, early exploration works have sought to use either \emph{sentiment-aware pre-training} \cite{Abdalla2019,Fu2018} or \emph{sentiment refinement} \cite{Utsumi2019,Yu2018,Yu2018b}. Both methods aim to obtain representations that are similar in semantics and sentiment. The \emph{sentiment-aware pre-training} accomplishes this goal by introducing external affective information as post-training supervision on word- or sentence-level, such as token sentiments and emoticons \cite{Zhou2020b}, aspect word \cite{Tian2020}, linguistic knowledge \cite{Ke2020}, and implicit sentiment information \cite{Li2021c}. The external sentiment knowledge can be introduced by performing word- or sentence-level polarity classification on the masked words \cite{Ke2020}. Nevertheless, learning word sentiment cannot help the model understand the sentiment intention of the whole sentence. Since the expressed sentiment of a sentence is not simply the sum of the polarities or the intensity of its constituent words.

\begin{figure*}
\centering
\subfloat[Unsupervised contrastive \\ learning]{\includegraphics[width=2.0in]{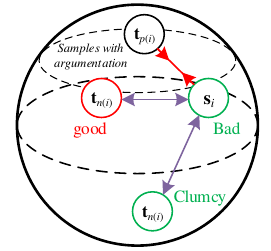}}
\subfloat[Supervised contrastive learning \\ with hard labels]{\includegraphics[width=2.0in]{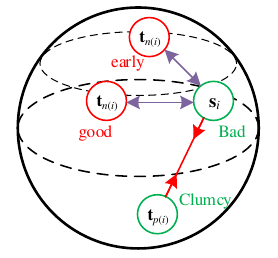}}
\subfloat[Supervised contrastive learning \\ with soft labels]{\includegraphics[width=2.0in]{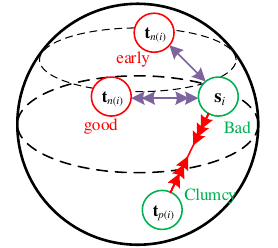}}
\caption{The conceptual diagram of using different contrastive learning strategies for sentiment-aware pre-training. (a) The self-supervised CL contrasts a single positive for each anchor (i.e., an augmentation of the anchor) against a set of negative consisting of the entire remainder of the batch. (b) The supervised CL contrasts the set of samples with same polarity as positives against the negatives from the remainder of the batch. (c) The proposed SoftMCL introducing external affective supervision to contrast the set of all samples according to the fine-grained distance of valence ratings between samples.}
\label{fig:1}
\end{figure*}

The \emph{sentiment refinement} model adjusts the vector of each affective word according to a given sentiment lexicon \cite{Utsumi2019,Yu2018,Yu2018b}. It can be used to adjust the representations of words so that they can be closer to semantically and sentimentally similar words and further away from sentimentally dissimilar ones. Such a method is applied as a post-processing step for already trained representations. An obvious risk is that the improper incorporation of affective information may excessively disperse the distribution of the semantic space.

Existing studies suggest introducing contrastive learning (CL) and using the same idea as \emph{sentiment refinement} to incorporate affective information. The idea involves training a model to pull together an anchor and a \emph{positive} sample in the latent space and push apart the anchor from many \emph{negative} samples. Semantic learning works \cite{Gao2021c} applied self-supervised CL, where a positive pair often consists of data augmentations of the sample, and negative pairs are formed by the anchor and randomly chosen samples in the batch. Unfortunately, this is unsuitable for sentiment-aware pre-training. If two words with the same sentiment polarity come from different samples in the same batch, they will be treated as negative sample pairs for CL, as the words \underline{\emph{clumsy}} and \underline{\emph{bad}} in Figure 1(a).

By leveraging sentiment information, recent studies \cite{Fan2022,Li2021c} applied supervised CL for sentiment-aware pre-training on both word- and sentence-level. The idea is transferred to that normalized representation from the words or sentences with the same sentiment polarities are pulled closer together than the ones with different sentiment polarities. By introducing external knowledge, such as sentiment-annotated lexicon or corpora, these models draw positives from samples with the same polarity as the anchor rather than data augmentation of the anchor, as shown in Figure 1(b).

A few practical issues should be considered when applying CL, especially for sentiment-aware pre-training. First is the batch size. As in most circumstances, larger batch size is better since that will provide more diverse and complicated negative samples, which is crucial for learning good representations \cite{Chen2022}. Nevertheless, the compatibility of the GPU memory often limits the batch size. Another important consideration is the quality of the negative samples. Hard negatives improve the representation quality learned by supervised CL if negative samples can be strictly classified according to the sentiment labels. It is evident that \underline{\emph{good}} and \underline{\emph{excellent}} are both positive, while the latter is stronger than the former. By defining only a few sentiment categories, e.g., positive, neutral, and negative, supervised CL can also cause all representations to converge to a few points, leading to the issue of latent space collapse.

This study proposed a soft momentum contrastive learning (SoftMCL) for fine-grained sentiment-aware pre-training. In contrast to hard labels of sentiment polarities, we introduce valence ratings \footnote{The valence ratings in Figure 1 are clumsy (4.14), bad (3.24), good (7.89), early (5.26)} from the extended version of affective norms of English words (E-ANEW) \cite{Warriner2013} and EmoBank \cite{Buechel2016,Buechel2017} as fine-grained soft-labels supervisions for CL. SoftMCL does not strictly distinguish between positive and negative samples, but measures the cross-entropy between the sentiment similarities in continuous affective space and distribution of semantic representations of these samples. Further, the proposed SoftMCL is conducted on both the word- and sentence-level to enhance the ability to learn affective information. A momentum queue was introduced to expand the contrastive samples, allowing storing and involving more negatives to overcome the limitations of hardware platforms.

Extensive experiments were conducted on four different sentiment-related tasks to evaluate the effectiveness of the proposed SoftMCL method. Empirical results show that SoftMCL performs better than other sentiment-aware pre-training methods.

The rest of the paper is organized as follows. Section 2 provides the preliminary knowledge of contrastive learning. Section 3 describes the proposed sentiment-aware momentum contrastive learning for fine-grained sentiment pre-training on both token- and sentence-level. Section 4 summarizes the comparative results against several baselines. Section 5 introduces the related works for affective state representation and sentiment pre-training. Conclusions are finally drawn in Section 6.

\section{Preliminary}

\subsection{Valence-Arousal Ratings}

Affective computing research offers two approaches to representing an emotional state: categorical and dimensional \cite{Calvo2013}. The categorical approach models affective states as several distinct classes, such as binary (positive and negative), Ekman's six fundamental emotions (anger, happiness, fear, sadness, disgust, and surprise) \cite{Ekman1992}, and Plutchik's eight fundamental emotions (which include Ekman's six as well as trust and anticipation) \cite{Plutchik1991}. A set of emotion categories must be established before classification. Custom-defined categories, however, typically cannot cover the complete emotional spectrum. As a result, it's possible to classify some affective states into an undefined category. Researchers may describe the same category under different names due to cultural or other differences, making it challenging to share categories. Based on this representation, several methods have been researched to provide practical applications, such as customer review analysis, mental illnesses identification, hotspot detection and forecasting, question answering, social text analysis, and financial market prediction.

These are the three fundamental limitations for discrete categorical encoding of sentiment:
 1) a set of emotion categories must be established before classification. Custom-defined categories, however, typically cannot cover the complete emotional spectrum. As a result, it's possible to classify some affective states into an undefined category.
 2) Research teams may describe the same category under different names due to cultural or other differences, making it challenging to share categories.
 3) The exceptional categories in particular domains must be re-defined when the field of application changes.

Continuous dimensional representations could be preferable because they express the affective state in a low-dimensional (2- or 3-dimensional) space. The most widely used method is based on Russell's valence-arousal space \cite{Russell1980}, which can precisely describe the affective state in a 2-dimensional continuous space \cite{Lee2022}. Valence and arousal are both real values. Valence is the pleasantness (both positive and negative) quality of a feeling. On the other side, arousal describes the intensity of the emotion's activation (such as excitement or calmness). This approach can represent any affective state as a point in the valence-arousal coordinate plane, effectively avoiding inconsistency and incompleteness in category definitions in a discrete representation and providing more intelligent and fine-grained sentiment applications \cite{Press2019,Wang2020a}.

\subsection{Contrastive Learning}

Contrastive learning is a self-supervised learning paradigm that captures inherent patterns and context in data without manual annotations \cite{Gao2021c}. The powerful representation learning ability of PLMs owes to the pre-training tasks in large-scale unlabeled data. Using contrastive learning, the PLMs can be trained by distinguishing the differences between samples. Given an anchor sample, a positive sample, and several negative samples, similarities are measured pairwise between the given anchor and the rest. Data augmentation usually produces a positive sample, meaning it belongs to the same class as the anchor sample. Conversely, negative samples indicate those not in the same class as the anchor. By measuring the similarity scores, the training objective of contrastive learning is,

\begin{equation}
{{\cal L}_{{\rm{CL}}}} =  - \sum\limits_{i \in I} {\log \frac{{\exp ({\bf{h}}_i^ \top  \cdot {{\bf{h}}_{p(i)}}/\tau )}}{{\sum\nolimits_{k \in A(i)} {\exp ({\bf{h}}_i^ \top  \cdot {{\bf{h}}_k}/\tau )} }}}
\label{Eq:1}
\end{equation}

\noindent where ${\bf{h}}_i$ denotes the hidden representation of the $i$th sample, $I$ is the set of samples in a batch, $\cdot $ is a dot product, $\tau $ is a temperature parameter used to control the density, $p(i)$ represents its positive sample, and $A(i)$ is the set that contains the positive sample and the negative samples.

\begin{figure*}[t!]
\centering
\includegraphics[width=5.8in]{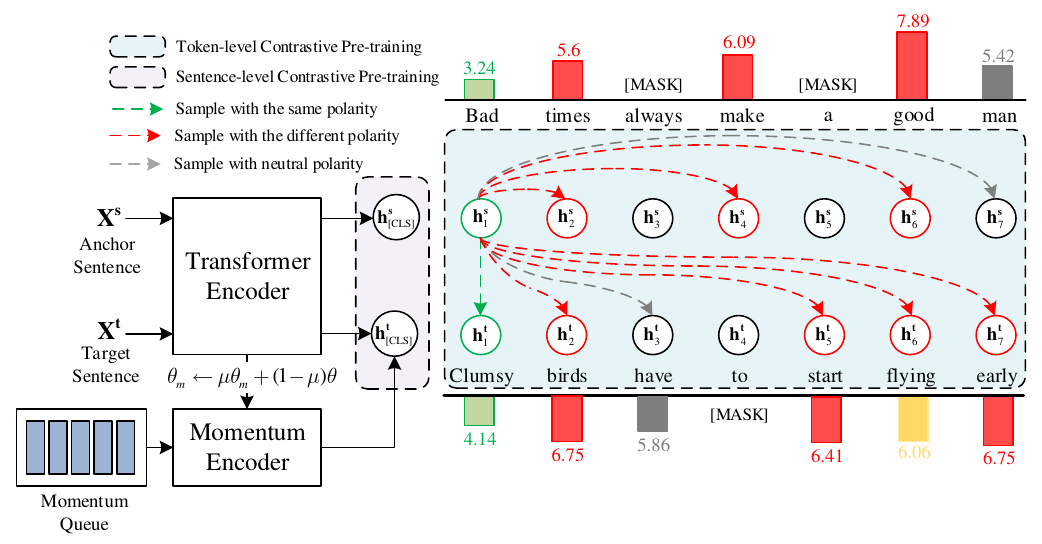}
\caption{ The overall architecture of the proposed SoftMCL. }
\label{fig:2}
\end{figure*}

\subsection{Supervised Contrastive Learning}

In addition to self-supervised learning, the prior knowledge of the labeled data can be incorporated. Since the anchor label is known, the label annotation can identify the positive and negative samples \cite{Fan2022,Li2021c}. Therefore, the training objective of contrastive learning can be generalized as follows,

\begin{equation}
\begin{adjustbox}{max width=\linewidth}
$
{{\cal L}_{{\rm{SCL}}}} =  - \sum\limits_{i \in I} {\sum\limits_{p \in P(i)} {\frac{1}{{\left| {P(i)} \right|}}\log \frac{{\exp ({\bf{h}}_i^ \top  \cdot {{\bf{h}}_p}/\tau )}}{{\sum\nolimits_{k \in A(i)} {\exp ({\bf{h}}_i^ \top  \cdot {{\bf{h}}_k}/\tau )} }}} }
$
\end{adjustbox}
\label{Eq:2}
\end{equation}

\noindent where $P(i)$ indicates all positives corresponding to anchor $i$ retrieved by the label annotation within a batch, and $|P(i)|$ is the number of positive samples. Using supervised learning, the PLMs are trained to pull the same class together while pushing apart clusters of samples from different classes.

In the objective of CL, the labels of positive (similar) pairs are set to 1, while the ones of negative (dissimilar) pairs are set to 0. Such hard labels lead to the issue of contrastive bias. Although two words have opposite emotional polarities, they may not differ significantly in valence.

\section{Soft Momentum Contrastive Learning}

Figure 2 shows the overall architecture of the proposed SoftMCL for fine-grained sentiment-aware pre-training. We introduce the affective ratings from the E-ANEW \cite{Warriner2013} and EmoBank \cite{Buechel2016,Buechel2017}. The difference in valence ratings between samples thus measures the sentiment similarity. The SoftMCL is performed on both token- and sentence-level to pull together the sentimentally similar samples but push away the dissimilar ones. In addition to the sentiment-aware pre-training, we introduce a momentum queue to expand the contrastive samples. The pre-trained encoder could be fine-tuned for the downstream applications.

\subsection{Sentiment Similarity}

All emotions can be annotated in continuous VA space as a real value in [1, 9] for valence, arousal, and dominance. Values 1, 5, and 9 denote valence's most negative, neutral, and positive affective states.

Given an input sentence ${\bf{X}} = \{ {x_1},{x_2},...,{x_n}\} $  where $n$ denotes the sequence length, each ${x_i} \in {\bf{X}}$ denotes the $i$th word. By adding a special token \texttt{[CLS]}, the input ${\bf{X}}$ is transformed into representation vectors, denoted as,

\begin{equation}
\begin{array}{l}
\{ {{\bf{h}}_{{\rm{[CLS]}}}},{{\bf{h}}_1},{{\bf{h}}_2},...,{{\bf{h}}_n}\} \\
\;\;\;\;\;\;\;\;\;\;\;\;\;\;\; = {\mathop{\rm Enc}\nolimits} (\{ {{\bf{x}}_{{\rm{[CLS]}}}},{{\bf{x}}_1},{{\bf{x}}_2},...,{{\bf{x}}_n}\} ;\theta )
\end{array}
\label{Eq:3}
\end{equation}

\noindent where ${\mathop{\rm Enc}\nolimits} ( \cdot )$ is the Transformer encoder parameterized by $\theta $. The associated affective ratings with both the sentence and words are obtained as, $Y = \{ {y_{{\rm{[CLS]}}}},{y_1},{y_2},...,{y_n}\} $  where ${y_{{\rm{[CLS]}}}}$ is the valence of the whole sentence, and the length $n$ in $Y$ should be equal to the length $n$ in ${\bf{X}}$. If an affective word was split to several tokens, all tokens should share the same valence as the word. Additionally, the $y_i$ is an annotated valence in a given affective lexicon. If a word $x_i$ does not appear in the affective lexicon, its corresponding $y_i$ will be assigned to 0, such that it will be masked during the word-level contrastive learning.

Based on the valence ratings, the sentiment similarity between two tokens or sentences can be defined as a normalized absolute similarity,

\begin{equation}
\Delta ({{\bf{h}}_1},{{\bf{h}}_2}) = 1 - \frac{{\left| {{y_1} - {y_2}} \right|}}{{{y_{\max }} - {y_{\min }}}}
\label{Eq:4}
\end{equation}

\noindent where $y_1$ and $y_2$ are the valence ratings of the tokens $x_1$ and $x_2$ or the sentences ${\bf{X}_1}$ and ${\bf{X}_2}$.

\subsection{Sentiment-aware Contrastive Learning}

The previous study applies unsupervised CL to contrast sentences within a batch. We perform both word- and sentence-level CL to enrich the sentiment information for the encoder. Using sentiment similarity of valence between samples, we do not strictly distinguish between positive and negative samples. Thus, the objective of sentiment-aware contrastive learning aims to measure the cross-entropy of the sentiment similarity distribution between samples and the similarity distribution of semantic representations, denoted as,

\begin{equation}
\begin{array}{l}
{{\cal L}_{{\rm{SentiCL}}}} = \\
- \sum\limits_{i \in I} {\sum\limits_{j \in B(i)} {\frac{{\Delta ({{\bf{h}}_i},{{\bf{h}}_j})}}{{\sum\nolimits_{l \in B(i)} {\Delta ({{\bf{h}}_i},{{\bf{h}}_l})} }}\log \frac{{\exp ({\bf{h}}_i^ \top  \cdot {{\bf{h}}_j}/\tau )}}{{\sum\nolimits_{k \in B(i)} {\exp ({\bf{h}}_i^ \top  \cdot {{\bf{h}}_k}/\tau )} }}} }
\end{array}
\label{Eq:5}
\end{equation}

\noindent where $B(i)$ is the batch size. For token-level CL, we randomly sampled 256 tokens belonging to the affective words appearing in the E-ANEW lexicon. For sentence-level CL, we extract hidden representation corresponding to \texttt{[CLS]} token, i.e., ${{\bf{h}}_{{\rm{[CLS]}}}}$ to measure the cross-entropy of the sentiment and semantic similarity distribution of the in-batch sentences.

\subsection{Momentum Contrastive Learning}

The batch size setting limits the number of contrastive samples. For sentence-level CL, each sample could only be paired with other $B(i)-1$ samples to create additional training patterns. To further expand the contrastive samples in batch, we proposed a momentum queue strategy, which uses the same idea as the MoCo \cite{He2020a} objective. We applied a momentum encoder ${{\mathop{\rm Enc}\nolimits} _m}( \cdot )$ to learn the sequence of hidden representations of the recent samples and store them within the affective ratings in a momentum queue ${\cal Q}$. These stored samples can expand the number of contrastive samples. For implementation, the momentum encoder ${{\mathop{\rm Enc}\nolimits} _m}( \cdot )$ is copied from the initial encoder ${\mathop{\rm Enc}\nolimits} ( \cdot )$. The parameters ${\theta _m}$ were updated from the parameter ${\theta}$ with a ratio $\mu $, denoted as,

\begin{equation}
{\theta _m} \leftarrow \mu {\theta _m} + (1 - \mu )\theta
\label{Eq:6}
\end{equation}

Based on the momentum queue ${\cal Q}$ and encoder ${{\mathop{\rm Enc}\nolimits} _m}( \cdot )$, there will be $Q_{(i)}$ samples that belong to ${\cal Q}$; thus, we obtain $B_{(i)}+Q_{(i)}$ contrastive samples. The loss function still measures cross-entropy between the distribution of sentiment similarities and the distribution of semantic representations,

\begin{equation}
\begin{adjustbox}{max width=\linewidth}
$
\begin{array}{l}
{{\cal L}_{{\rm{MCL}}}} = \\
 - \sum\limits_{i \in I} {\left[ {\sum\limits_{j \in B(i)} {\frac{{\Delta ({{\bf{h}}_i},{{\bf{h}}_j})}}{{\sum\nolimits_{l \in \{ B(i),Q(i)\} } {\Delta ({{\bf{h}}_i},{{\bf{h}}_l})} }}\log \frac{{\exp ({\bf{h}}_i^ \top  \cdot {{\bf{h}}_j}/\tau )}}{{\sum\nolimits_{k \in \{ B(i),Q(i)\} } {\exp ({\bf{h}}_i^ \top  \cdot {{\bf{h}}_k}/\tau )} }}} } \right.} \\
\left. { + \sum\limits_{m \in Q(i)} {\frac{{\Delta ({{\bf{h}}_i},{{\bf{h}}_m})}}{{\sum\nolimits_{l \in \{ B(i),Q(i)\} } {\Delta ({{\bf{h}}_i},{{\bf{h}}_l})} }}\log \frac{{\exp ({\bf{h}}_i^ \top  \cdot {{\bf{h}}_m}/\tau )}}{{\sum\nolimits_{k \in \{ B(i),Q(i)\} } {\exp ({\bf{h}}_i^ \top  \cdot {{\bf{h}}_k}/\tau )} }}} } \right]
\end{array}
$
\end{adjustbox}
\label{Eq:7}
\end{equation}

\subsection{Training Objectives}

In addition to sentiment-aware pre-training, masked language modeling (MLM) was also implemented for pre-training. Like the previous settings, 15\% of words were first masked. Further, we masked affective words in the input sentence using the E-ANEW lexicon with a specific ratio  . The MLM objective is to predict the masked words within the given contexts. Then, SoftMCL was performed to distinguish the valence difference between samples. The training objective is,

\begin{equation}
\label{Eq:8}
{\cal L} = {{\cal L}_{MLM}} + {\lambda _1}{\cal L}_{{\mathop{\rm MCL}\nolimits} }^{word} + {\lambda _2}{\cal L}_{{\mathop{\rm MCL}\nolimits} }^{sent}
\end{equation}

\noindent where ${\lambda _1}$ and ${\lambda _2}$ balance the contribution of different subtasks.

\begin{table}[t!]
\centering
\begin{adjustbox}{max width=\linewidth}
$
\begin{tabular}{|l|c|c|c|c|c|}
\hline
\multicolumn{1}{|c|}{\textbf{Dataset}} & \textbf{Train} & \textbf{Dev} & \textbf{Test} & \textbf{Length} & \textbf{\#C} \\ \hline
SST-2                                  & 8,544          & 1,101        & 2,210         & 19.2            & 2            \\ \hline
SST-5                                  & 6,920          & 872          & 1,821         & 19.2            & 5            \\ \hline
MR                                     & 8,534          & 1,078        & 1,050         & 21.7            & 2            \\ \hline
CR                                     & 2,025          & 675          & 675           & 20.1            & 2            \\ \hline
IMDB                                   & 22,500         & 2,500        & 25,000        & 279.2           & 2            \\ \hline
Yelp-2                                 & 504,000        & 56,000       & 38,000        & 155.3           & 2            \\ \hline
Yelp-5                                 & 594,000        & 56,000       & 50,000        & 156.6           & 5            \\ \hline
Emobank                                & 6,195          & 2,065        & 2,065         & 17.6            & -            \\ \hline
Facebook                               & 1,736          & 579          & 579           & 19.8            & -            \\ \hline
Lap14                                  & 3,452          & 150          & 676           & 30.2            & 3            \\ \hline
Rest14                                 & 2,163          & 150          & 638           & 25.6            & 3            \\ \hline
\end{tabular}
$
\end{adjustbox}
\caption{Statistics of datasets used in the experiments. \#C indicates the number of the labels.}
\label{tab:1}
\end{table}

For fine-tuning, the pre-trained model generates a sequence of hidden representations using Eq.(3). The aspect term was appended to the sequence with the \texttt{[SEP]} token for aspect-level sentiment classification. The hidden representation of the \texttt{[CLS]} token was used to predict the sentiment intensity or polarity of the input sample.

\section{Experiments}

\subsection{Tasks and Datasets}

\textbf{Sentiment-aware Pre-training.} The proposed SoftMCL was pre-trained using the same dataset as RoBERTa \cite{Liu2019a} and DeBERTa \cite{He2021}, which is a combination of Wikipedia, Bookcorpus, CCNews, Stories, and OpenWebText. We sampled 2 million sentences with a maximum length of 128 for word-level pre-training. We introduce the E-ANEW \cite{Warriner2013} to provide the word-level VA annotation. For sentence-level CL, we adopted a corpus from the training split of EmoBank \cite{Buechel2016,Buechel2017}, which contains 40,000 sentences with valence annotation.

The empirical experiments were conducted on four sentiment-related tasks: phrase-level intensity prediction, sentence-level classification and regression, and aspect-level sentiment analysis. The details are presented as follows.

\begin{table}[t!]
\footnotesize
\centering
\begin{tabular}{|lcccc|}
\hline
\multicolumn{1}{|c|}{\multirow{2}{*}{\textbf{Model}}} & \multicolumn{2}{c|}{\textbf{Word}}                                & \multicolumn{2}{c|}{\textbf{Phrase}}                              \\ \cline{2-5}
\multicolumn{1}{|c|}{}                       & \multicolumn{1}{c|}{$k \uparrow$}   & \multicolumn{1}{c|}{$\rho \uparrow$} & \multicolumn{1}{c|}{$k \uparrow$}   & \multicolumn{1}{c|}{$\rho \uparrow$} \\ \hline
\multicolumn{5}{|c|}{General pre-trained models}                                                                                                                   \\ \hline
\multicolumn{1}{|l|}{BERT}                   & \multicolumn{1}{l|}{0.638} & \multicolumn{1}{l|}{0.841}  & \multicolumn{1}{l|}{0.624} & 0.829                       \\ \hline
\multicolumn{1}{|l|}{XLNet}                  & \multicolumn{1}{l|}{0.644} & \multicolumn{1}{l|}{0.852}  & \multicolumn{1}{l|}{0.637} & 0.833                       \\ \hline
\multicolumn{1}{|l|}{RoBERTa}                & \multicolumn{1}{l|}{0.656} & \multicolumn{1}{l|}{0.868}  & \multicolumn{1}{l|}{0.642} & 0.849                       \\ \hline
\multicolumn{1}{|l|}{DeBERTa}                & \multicolumn{1}{l|}{0.667} & \multicolumn{1}{l|}{0.866}  & \multicolumn{1}{l|}{0.644} & 0.852                       \\ \hline
\multicolumn{5}{|c|}{Sentiment-aware pre-trained models}                                                                                                           \\ \hline
\multicolumn{1}{|l|}{BERT-PT}                & \multicolumn{1}{l|}{0.704} & \multicolumn{1}{l|}{0.880}  & \multicolumn{1}{l|}{0.688} & 0.877                       \\ \hline
\multicolumn{1}{|l|}{SentiBERT}              & \multicolumn{1}{l|}{0.699} & \multicolumn{1}{l|}{0.886}  & \multicolumn{1}{l|}{0.692} & 0.882                       \\ \hline
\multicolumn{1}{|l|}{SentiLARE}              & \multicolumn{1}{l|}{0.712} & \multicolumn{1}{l|}{0.894}  & \multicolumn{1}{l|}{0.702} & 0.891                       \\ \hline
\multicolumn{5}{|c|}{Proposed models with ablation study}                                                                                                          \\ \hline
\multicolumn{1}{|l|}{SoftMCL}                & \multicolumn{1}{l|}{\textbf{0.778}} & \multicolumn{1}{l|}{\textbf{0.928}}  & \multicolumn{1}{l|}{\textbf{0.775}} & \textbf{0.922}                      \\ \hline
\multicolumn{1}{|l|}{\quad w/o WP}                 & \multicolumn{1}{l|}{0.698} & \multicolumn{1}{l|}{0.878}  & \multicolumn{1}{l|}{0.692} & 0.874                       \\ \hline
\multicolumn{1}{|l|}{\quad w/o SP}                 & \multicolumn{1}{l|}{0.702} & \multicolumn{1}{l|}{0.889}  & \multicolumn{1}{l|}{0.685} & 0.865                       \\ \hline
\multicolumn{1}{|l|}{\quad w/o MoCL}               & \multicolumn{1}{l|}{0.724} & \multicolumn{1}{l|}{0.895}  & \multicolumn{1}{l|}{0.713} & 0.904                       \\ \hline
\end{tabular}
\caption{Comparative results of different models on phrase-level intensity prediction task.}
\label{tab:2}
\end{table}

\begin{table*}[t!]
\begin{adjustbox}{max width=\linewidth}
$
\centering
\begin{tabular}{lccccccccccc}
\hline
\multicolumn{1}{|c|}{\multirow{3}{*}{\textbf{Model}}} & \multicolumn{7}{c|}{\textbf{Sentence-level sentiment classification}}                                                                                                                                                                                                & \multicolumn{4}{c|}{\textbf{Aspect-level sentiment classification}}                                                    \\ \cline{2-12}
\multicolumn{1}{|c|}{}                                & \multicolumn{1}{c|}{\textbf{SST-2}} & \multicolumn{1}{c|}{\textbf{SST-5}} & \multicolumn{1}{c|}{\textbf{MR}}  & \multicolumn{1}{c|}{\textbf{CR}}  & \multicolumn{1}{c|}{\textbf{IMDB}} & \multicolumn{1}{c|}{\textbf{Yelp-2}} & \multicolumn{1}{c|}{\textbf{Yelp-5}} & \multicolumn{2}{c|}{\textbf{Lap14}}                                  & \multicolumn{2}{c|}{\textbf{Rest14}}            \\ \cline{2-12}
\multicolumn{1}{|c|}{}                                & \multicolumn{1}{c|}{\textbf{Acc} $\uparrow$}   & \multicolumn{1}{c|}{\textbf{Acc} $\uparrow$}   & \multicolumn{1}{c|}{\textbf{Acc} $\uparrow$} & \multicolumn{1}{c|}{\textbf{Acc} $\uparrow$} & \multicolumn{1}{c|}{\textbf{Acc} $\uparrow$}  & \multicolumn{1}{c|}{\textbf{Acc} $\uparrow$}    & \multicolumn{1}{c|}{\textbf{Acc}}    & \multicolumn{1}{c|}{\textbf{Acc} $\uparrow$} & \multicolumn{1}{c|}{\textbf{$F_1$-score} $\uparrow$} & \multicolumn{1}{c|}{\textbf{Acc} $\uparrow$} & \multicolumn{1}{c|}{\textbf{$F_1$-score} $\uparrow$} \\ \hline
\multicolumn{12}{|c|}{General pretrained models}                                                                                                                                                                                                                                                                                                                                                                                                      \\ \hline
\multicolumn{1}{|l|}{BERT}                            & \multicolumn{1}{c|}{90.82}          & \multicolumn{1}{c|}{53.37}          & \multicolumn{1}{c|}{87.52}        & \multicolumn{1}{c|}{88.22}        & \multicolumn{1}{c|}{93.87}         & \multicolumn{1}{c|}{97.74}           & \multicolumn{1}{c|}{70.16}           & \multicolumn{1}{c|}{78.53}        & \multicolumn{1}{c|}{73.11}       & \multicolumn{1}{c|}{83.77}        & \multicolumn{1}{c|}{76.06}      \\ \hline
\multicolumn{1}{|l|}{XLNet}                           & \multicolumn{1}{c|}{92.04}          & \multicolumn{1}{c|}{56.33}          & \multicolumn{1}{c|}{89.45}        & \multicolumn{1}{c|}{88.33}        & \multicolumn{1}{c|}{96.21}         & \multicolumn{1}{c|}{97.41}           & \multicolumn{1}{c|}{70.23}           & \multicolumn{1}{c|}{80.00}        & \multicolumn{1}{c|}{75.88}       & \multicolumn{1}{c|}{84.93}        & \multicolumn{1}{c|}{76.70}       \\ \hline
\multicolumn{1}{|l|}{RoBERTa}                         & \multicolumn{1}{c|}{92.28}          & \multicolumn{1}{c|}{54.89}          & \multicolumn{1}{c|}{89.41}        & \multicolumn{1}{c|}{88.47}        & \multicolumn{1}{c|}{94.68}         & \multicolumn{1}{c|}{97.98}           & \multicolumn{1}{c|}{70.12}           & \multicolumn{1}{c|}{81.03}        & \multicolumn{1}{c|}{77.16}       & \multicolumn{1}{c|}{86.07}        & \multicolumn{1}{c|}{79.21}       \\ \hline
\multicolumn{1}{|l|}{DeBERTa}                         & \multicolumn{1}{c|}{94.82}          & \multicolumn{1}{c|}{56.89}          & \multicolumn{1}{c|}{89.52}        & \multicolumn{1}{c|}{88.49}        & \multicolumn{1}{c|}{94.92}         & \multicolumn{1}{c|}{97.88}           & \multicolumn{1}{c|}{70.22}           & \multicolumn{1}{c|}{81.14}        & \multicolumn{1}{c|}{77.22}       & \multicolumn{1}{c|}{86.12}        & \multicolumn{1}{c|}{79.43}      \\ \hline
\multicolumn{12}{|c|}{Sentiment-aware pre-trained models}                                                                                                                                                                                                                                                                                                                                                                                             \\ \hline
\multicolumn{1}{|l|}{BERT-PT}                         & \multicolumn{1}{c|}{91.42}          & \multicolumn{1}{c|}{53.24}          & \multicolumn{1}{c|}{87.30}        & \multicolumn{1}{c|}{86.52}        & \multicolumn{1}{c|}{93.99}         & \multicolumn{1}{c|}{97.77}           & \multicolumn{1}{c|}{69.90}           & \multicolumn{1}{c|}{78.46}        & \multicolumn{1}{c|}{73.82}       & \multicolumn{1}{c|}{85.58}        & \multicolumn{1}{c|}{77.99}      \\ \hline
\multicolumn{1}{|l|}{SentiBERT}                       & \multicolumn{1}{c|}{92.18}          & \multicolumn{1}{c|}{56.87}          & \multicolumn{1}{c|}{88.59}        & \multicolumn{1}{c|}{88.24}        & \multicolumn{1}{c|}{94.04}         & \multicolumn{1}{c|}{97.66}           & \multicolumn{1}{c|}{69.94}           & \multicolumn{1}{c|}{76.87}        & \multicolumn{1}{c|}{71.74}       & \multicolumn{1}{c|}{83.71}        & \multicolumn{1}{c|}{75.42}       \\ \hline
\multicolumn{1}{|l|}{SentiLARE}                       & \multicolumn{1}{c|}{94.28}          & \multicolumn{1}{c|}{58.59}          & \multicolumn{1}{c|}{90.82}        & \multicolumn{1}{c|}{89.52}        & \multicolumn{1}{c|}{95.71}         & \multicolumn{1}{c|}{98.22}           & \multicolumn{1}{c|}{71.57}           & \multicolumn{1}{c|}{82.16}        & \multicolumn{1}{c|}{78.70}       & \multicolumn{1}{c|}{88.32}        & \multicolumn{1}{c|}{81.63}       \\ \hline
\multicolumn{1}{|l|}{SENTIX}                          & \multicolumn{1}{c|}{93.30}          & \multicolumn{1}{c|}{55.57}          & \multicolumn{1}{c|}{-}            & \multicolumn{1}{c|}{88.72}        & \multicolumn{1}{c|}{94.78}         & \multicolumn{1}{c|}{97.83}           & \multicolumn{1}{c|}{-}               & \multicolumn{1}{c|}{80.56}        & \multicolumn{1}{c|}{-}           & \multicolumn{1}{c|}{87.32}        & \multicolumn{1}{c|}{-}          \\ \hline
\multicolumn{1}{|l|}{SCAPT}                           & \multicolumn{1}{c|}{-}              & \multicolumn{1}{c|}{-}              & \multicolumn{1}{c|}{-}            & \multicolumn{1}{c|}{-}            & \multicolumn{1}{c|}{-}             & \multicolumn{1}{c|}{-}               & \multicolumn{1}{c|}{-}               & \multicolumn{1}{c|}{82.76}        & \multicolumn{1}{c|}{79.15}       & \multicolumn{1}{c|}{89.11}        & \multicolumn{1}{c|}{83.91}       \\ \hline
\multicolumn{12}{|c|}{Proposed model with ablation study}                                                                                                                                                                                                                                                                                                                                                                                             \\ \hline
\multicolumn{1}{|l|}{SoftMCL}                         & \multicolumn{1}{c|}{\textbf{96.54}}         & \multicolumn{1}{c|}{\textbf{59.82}}         & \multicolumn{1}{c|}{\textbf{92.56}}        & \multicolumn{1}{c|}{\textbf{90.02}}        & \multicolumn{1}{c|}{\textbf{96.85}}         & \multicolumn{1}{c|}{\textbf{98.29}}           & \multicolumn{1}{c|}{\textbf{71.76}}           & \multicolumn{1}{c|}{\textbf{83.82}}        & \multicolumn{1}{c|}{\textbf{81.08}}       & \multicolumn{1}{c|}{\textbf{89.23}}        & \multicolumn{1}{c|}{\textbf{83.95}}       \\ \hline
\multicolumn{1}{|l|}{\quad w/o WP}                          & \multicolumn{1}{c|}{94.01}          & \multicolumn{1}{c|}{58.25}          & \multicolumn{1}{c|}{90.14}        & \multicolumn{1}{c|}{87.67}        & \multicolumn{1}{c|}{94.31}         & \multicolumn{1}{c|}{95.73}           & \multicolumn{1}{c|}{69.88}           & \multicolumn{1}{c|}{81.63}        & \multicolumn{1}{c|}{78.96}       & \multicolumn{1}{c|}{86.91}        & \multicolumn{1}{c|}{81.75}       \\ \hline
\multicolumn{1}{|l|}{\quad w/o SP}                          & \multicolumn{1}{c|}{93.65}          & \multicolumn{1}{c|}{58.04}          & \multicolumn{1}{c|}{89.79}        & \multicolumn{1}{c|}{87.32}        & \multicolumn{1}{c|}{93.96}         & \multicolumn{1}{c|}{95.35}           & \multicolumn{1}{c|}{69.62}           & \multicolumn{1}{c|}{81.31}        & \multicolumn{1}{c|}{78.66}       & \multicolumn{1}{c|}{86.57}        & \multicolumn{1}{c|}{81.43}       \\ \hline
\multicolumn{1}{|l|}{\quad w/o MoCL}                        & \multicolumn{1}{c|}{91.33}          & \multicolumn{1}{c|}{56.61}          & \multicolumn{1}{c|}{87.55}        & \multicolumn{1}{c|}{85.15}        & \multicolumn{1}{c|}{91.62}         & \multicolumn{1}{c|}{92.99}           & \multicolumn{1}{c|}{67.89}           & \multicolumn{1}{c|}{79.29}        & \multicolumn{1}{c|}{76.7}        & \multicolumn{1}{c|}{84.42}        & \multicolumn{1}{c|}{79.41}       \\ \hline
\end{tabular}
$
\end{adjustbox}
\caption{Comparative results of different models on sentence-level and aspect-level sentiment classification tasks.}
\label{tab:3}
\end{table*}

\begin{table}[t!]
\footnotesize
\centering
\begin{tabular}{|lcccc|}
\hline
\multicolumn{1}{|c|}{\multirow{2}{*}{\textbf{Model}}} & \multicolumn{2}{c|}{\textbf{EmoBank}}                          & \multicolumn{2}{c|}{\textbf{Facebook}}    \\ \cline{2-5}
\multicolumn{1}{|c|}{}                                & \multicolumn{1}{c|}{\textbf{MAE} $\downarrow$} & \multicolumn{1}{c|}{$r \uparrow$}   & \multicolumn{1}{c|}{\textbf{MAE} $\downarrow$} & $r \uparrow$   \\ \hline
\multicolumn{5}{|c|}{General pre-trained models}                                                                                                                   \\ \hline
\multicolumn{1}{|l|}{BERT}                            & \multicolumn{1}{c|}{0.581}        & \multicolumn{1}{c|}{0.521} & \multicolumn{1}{c|}{0.719}        & 0.635 \\ \hline
\multicolumn{1}{|l|}{XLNet}                           & \multicolumn{1}{c|}{0.523}        & \multicolumn{1}{c|}{0.589} & \multicolumn{1}{c|}{0.706}        & 0.654 \\ \hline
\multicolumn{1}{|l|}{RoBERTa}                         & \multicolumn{1}{c|}{0.518}        & \multicolumn{1}{c|}{0.592} & \multicolumn{1}{c|}{0.694}        & 0.642 \\ \hline
\multicolumn{1}{|l|}{DeBERTa}                         & \multicolumn{1}{c|}{0.514}        & \multicolumn{1}{c|}{0.591} & \multicolumn{1}{c|}{0.711}        & 0.631 \\ \hline
\multicolumn{5}{|c|}{Sentiment-aware pre-trained models}                                                                                                           \\ \hline
\multicolumn{1}{|l|}{BERT-PT}                         & \multicolumn{1}{c|}{0.506}        & \multicolumn{1}{c|}{0.610} & \multicolumn{1}{c|}{0.708}        & 0.631 \\ \hline
\multicolumn{1}{|l|}{SentiBERT}                       & \multicolumn{1}{c|}{0.505}        & \multicolumn{1}{c|}{0.612} & \multicolumn{1}{c|}{0.688}        & 0.677 \\ \hline
\multicolumn{1}{|l|}{SentiLARE}                       & \multicolumn{1}{c|}{0.498}        & \multicolumn{1}{c|}{0.615} & \multicolumn{1}{c|}{0.669}        & 0.671 \\ \hline
\multicolumn{5}{|c|}{Proposed model with ablation study}                                                                                                           \\ \hline
\multicolumn{1}{|l|}{SoftMCL}                         & \multicolumn{1}{c|}{\textbf{0.462}}        & \multicolumn{1}{c|}{\textbf{0.639}} & \multicolumn{1}{c|}{\textbf{0.642}}        & \textbf{0.685} \\ \hline
\multicolumn{1}{|l|}{\quad w/o WP}                          & \multicolumn{1}{c|}{0.475}        & \multicolumn{1}{c|}{0.629} & \multicolumn{1}{c|}{0.663}        & 0.661 \\ \hline
\multicolumn{1}{|l|}{\quad w/o SP}                          & \multicolumn{1}{c|}{0.495}        & \multicolumn{1}{c|}{0.620} & \multicolumn{1}{c|}{0.734}        & 0.634 \\ \hline
\multicolumn{1}{|l|}{\quad w/o MoCL}                        & \multicolumn{1}{c|}{0.492}        & \multicolumn{1}{c|}{0.622} & \multicolumn{1}{c|}{0.710}        & 0.662 \\ \hline
\end{tabular}
\caption{Comparative results of different models on sentence-level regression task to predict valence ratings.}
\label{tab:4}
\end{table}

\noindent \textbf{Phrase-level Intensity Prediction.} The dataset was the SemEval-2016 task 7 (general English subtask) \cite{Kiritchenko2016}, which contained 2,799 terms, including 1,330 \textbf{words} and 1,469 \textbf{phrases}. Each term was annotated within the range of [0, 1]. The metrics contain Kendall's rank correlation coefficient ($k$) and Spearman's rank correlation coefficient ($\rho $).

\noindent \textbf{Sentence-level Classification.} The benchmarks include Stanford Sentiment Treebank (\textbf{SST-2/5}) \cite{Socher2013b}, Movie Review (\textbf{MR}) \cite{Pang2005}, and Customer Review (\textbf{CR}) \cite{Conneau2019}, \textbf{IMDB} \cite{Maas2011}  and \textbf{Yelp-2/5} \cite{Zhang2013a}. The evaluation metric is accuracy.

\noindent \textbf{Sentence-level Regression.} Instead of using binary or fine-grained labels, both the test split of \textbf{Emobank} \cite{Buechel2016,Buechel2017} and \textbf{Facebook} \cite{Preotiuc-Pietro2016} use real-valued VA ratings in the range of (1, 9). The valence denotes the degree of positive and negative sentiment. The metrics are mean absolute error (MAE) and Pearson's correlation coefficient ($r$).

\noindent \textbf{Aspect-level Sentiment Classification.} This task requires the identification of aspects in text, as well as the assignment of sentiment labels to those aspects. The datasets are SemEval-2014 task 4 \cite{Pontiki2014} in laptop (\textbf{Lap14}) and restaurant (\textbf{Res14}). The evaluation metric is accuracy and $F_1$-score.

Table \ref{tab:1} summarizes the details of these datasets, including the amount of training, validation, and test splits. For MR, IMDB, and Yelp-2/5, a subset was randomly selected from the training set for validation. For CR, Emobank, and Facebook, the evaluation was performed using a 5-fold cross-validation. For each round, each dataset was randomly split into a training, development, and test set with a 6:2:2 ratio.

\subsection{Baselines}

The proposed SoftMCL was compared with several pre-trained models, pre-trained by general tasks, i.e., MLM and NSP, and sentiment-related tasks.

The general-purpose pre-trained models include BERT \cite{Devlin2018}, XLNet \cite{Yang2019a}, RoBERTa \cite{Liu2019a}, and DeBERTa \cite{He2021}. The sentiment-aware pretrained models include BERT-PT \cite{Xu2019b}, SentiBERT \cite{Yin2020}, SentiLARE \cite{Ke2020}, SENTIX \cite{Zhou2020b}, and SCAPT \cite{Li2021c}.

For phrase-level intensity prediction, we further report the empirical results from the original study with the same implementation settings \cite{Yu2018,Yu2018b}.

\subsection{Implementation Details}

The backbone model we used is DeBERTa \cite{He2021}, which is initialized from the pre-trained checkpoint of the base version  using the HuggingFace toolkit \cite{Wolf2020}. We use the AdamW optimizer \cite{Kingma2014} with a linear learning rate scheduler with 10\% warm-up steps for pre-training. The learning rate is initialized with 2e-5. The maximum length is truncated to 128. The batch size of contrastive learning is 64. The steps of warm-up and pre-training are 1,500 and 20,000, respectively.

For fine-tuning, the hyper-parameters were selected by using a grid search strategy. The number of epochs for each model was set depending on an early strategy, where the patient was set to 3. The model exits when the loss does not decrease in three epochs.

\subsection{Comparative Results}

This section compared the proposed SoftMCL against several previous methods on four sentiment-related tasks. Tables 2, 3, and 4 summarize the empirical results of the proposed SoftMCL against different baselines on four different tasks. As indicated, the proposed SoftMCL consistently outperformed all baselines for different tasks, demonstrating the superiority of the proposed SoftMCL in capturing affective information in addition to semantic learning.

Moreover, the proposed SoftMCL performs significantly better than DeBERTa on all datasets for four different sentiment-related tasks, indicating the effectiveness of the proposed momentum contrastive learning method. Compared with other sentiment-aware pre-trained methods, the improvement mainly comes from twofolds. First, the valence ratings are introduced as soft labels to supervise contrastive learning. Further, the CL is performed on both word- and sentence-level to guide the model to learn more fine-grained sentiment features.

\subsection{Ablation Study}

\begin{figure}[t!]
\centering
\includegraphics[width=3.0in]{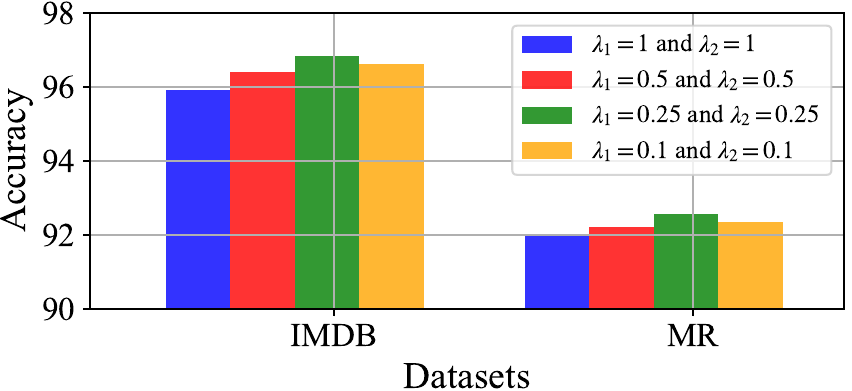}
\caption{  The effect of different balance coefficient. }
\label{fig:3}
\end{figure}

\begin{figure}[t!]
\centering
\subfloat[IMDB]{\includegraphics[width=1.5in]{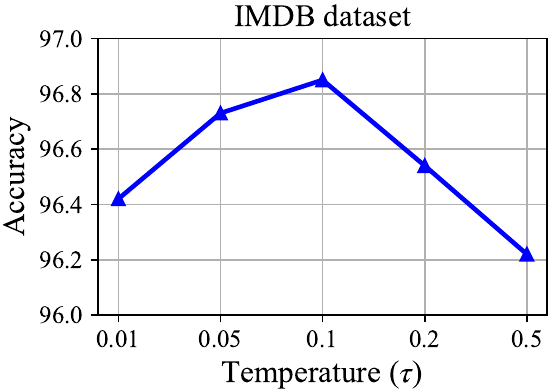}}
\subfloat[MR]{\includegraphics[width=1.5in]{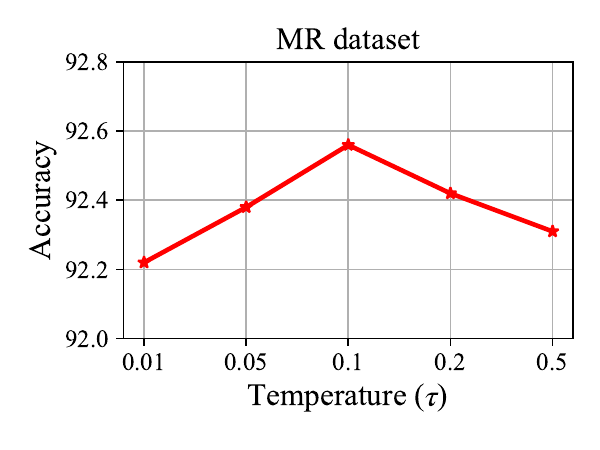}}
\caption{The effect of different temperature.}
\label{fig:4}
\end{figure}

\begin{figure}[t!]
\centering
\subfloat[IMDB]{\includegraphics[width=1.5in]{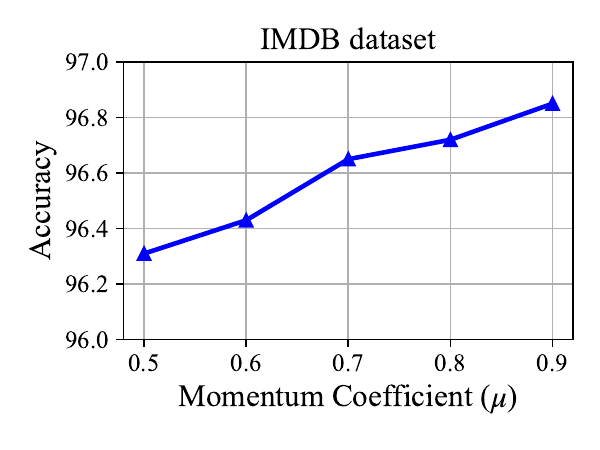}}
\subfloat[MR]{\includegraphics[width=1.5in]{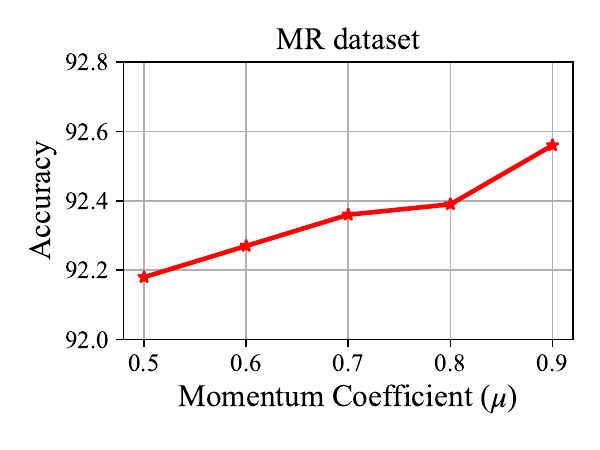}}
\caption{The effect of different momentum coefficient.}
\label{fig:5}
\end{figure}

\begin{figure}[t!]
\centering
\subfloat[IMDB]{\includegraphics[width=1.5in]{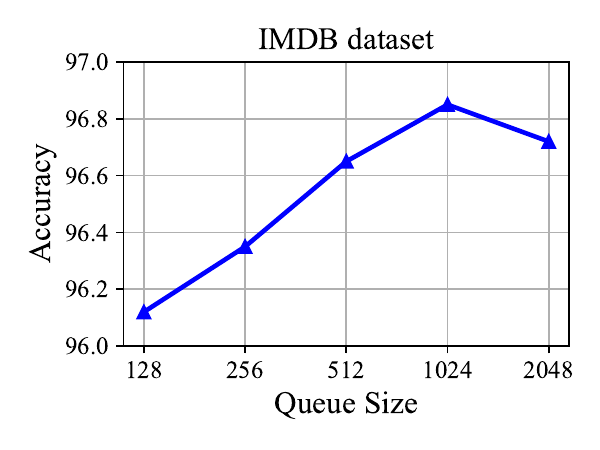}}
\subfloat[MR]{\includegraphics[width=1.5in]{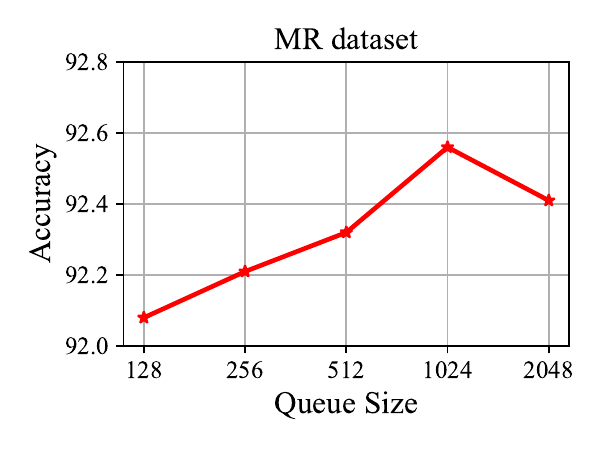}}
\caption{The effect of different queue size.}
\label{fig:6}
\end{figure}

\begin{figure*}[t!]
\centering
\subfloat[SoftMCL word-level pre-training]{\includegraphics[width=2.8in]{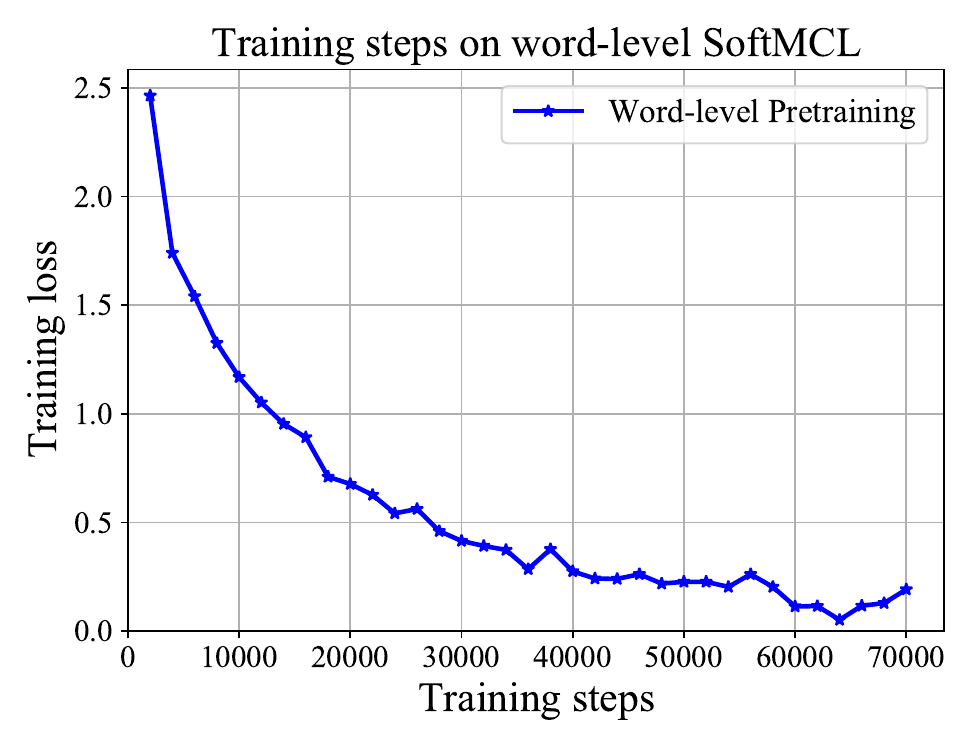}}
\subfloat[SoftMCL sentence-level pretraining]{\includegraphics[width=2.8in]{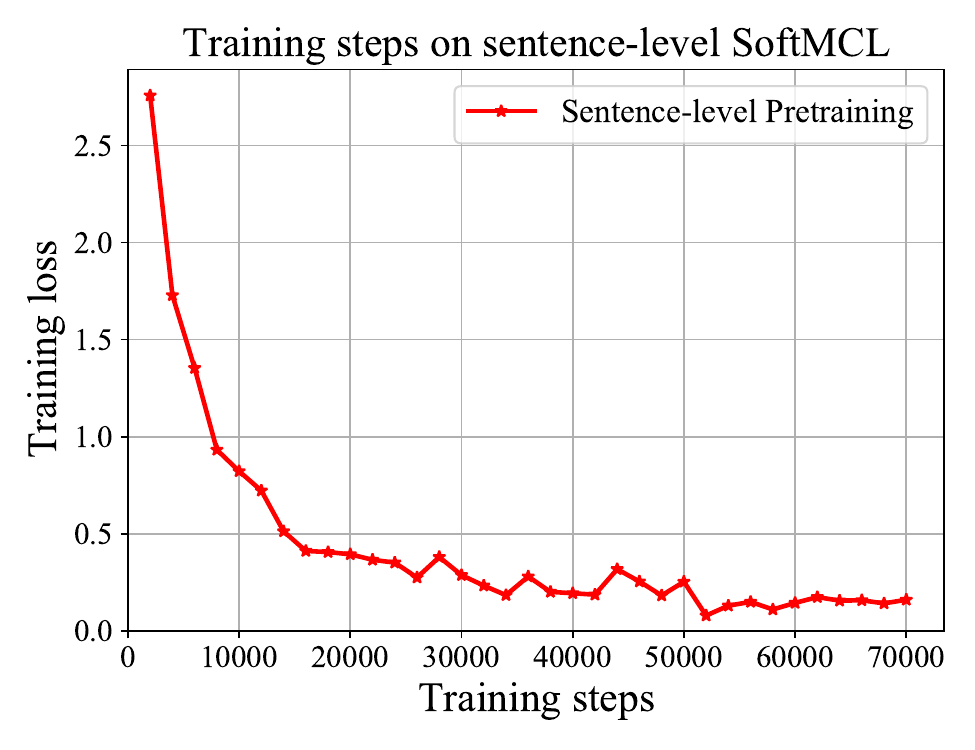}}
\caption{Training curve of the proposed SoftMCL on word- and sentence-level pretraining.}
\label{fig:7}
\end{figure*}

We successively removed word-level pre-training (WP), sentence-level pre-training (SP), and the momentum CL (MoCL) to investigate whether it would reduce performance. The bottom rows of Tables 2, 3, and 4 show the ablation results, indicating each removal produces a varying degree of performance decline. This also proves the necessity of tailoring pre-training paradigms for sentiment-related tasks. The word- and sentence-level pre-training tasks capture affective information in different granularity, and combining multi-granularity pre-training benefits performance improvement.

Further, SoftMCL w/o sentence-level pre-training is worse than w/o word-level pre-training in most circumstances. The performance decline can be consistently observed on aspect-level sentiment classification tasks. The rationale is that the global context is essential for analyzing aspect sentiments while focusing only on words leads to less robust prediction.

By removing the momentum queue and encoder, the proposed SoftMCL will degenerate to only using contrastive samples from the same batch for CL. This will decrease the number of contrastive samples, thus limiting the performance of the proposed SoftMCL.

\subsection{Impacts of Parameters}

\noindent \textbf{Balance Coefficient.} The hyperparameters ${\lambda _1}$ and ${\lambda _2}$ in Eq. (8) were applied to control the contribution of word- and sentence-level CL to the total training. Figure 3 investigates the effect of different combinations of these two parameters. The observation is that the MLM loss is still essential for the pre-training. When ${\lambda _1}=0.25$ and ${\lambda _2}=0.25$, the SoftMCL can achieve the best results on IMDB and MR datasets.

\noindent \textbf{Temperature.} The temperature $\tau $ in the training loss of CL controls the density of the softmax function. Figure 4 shows the effect of temperature on IMDB and MR of the SoftMCL. As indicated, all the best results can be achieved using a temperature $\tau=0.1$. Smaller temperatures benefit training more than higher ones, but extremely low temperatures are more challenging for CL due to numerical instability.

\noindent \textbf{Momentum Coefficient.} The momentum coefficient $\mu $ can be used to update the parameters of the momentum encoder, making the momentum encoder evolve more smoothly than the original encoder. Figure 5 shows the effect of the momentum coefficient  on IMDB and MR datasets. As indicated, a relatively large momentum, e.g., $\mu=0.9$ works much better than a smaller one, suggesting that a slowly evolving original encoder is essential to utilize the momentum queue.

\noindent \textbf{Queue Size.} The size of the momentum queue determines the number of contrastive samples that can be stored. Figure 6 evaluates the impact of queue size on the performance of the SoftMCL. As indicated, the best performance can be achieved when the size is 1024. A large queue requires better hardware but introduces more noise samples.

\subsection{Training Analysis}

Figure 7 shows the training loss curve of the proposed SoftMCL on word- and sentence-level pre-training. The SoftMCL model gradually converges within 40,000 steps for word-level pre-training since enough emotional tokens can be introduced from each batch for word-level CL training. Even if the batch size setting for sentence-level pre-training limits the number of samples used for CL in a batch, the SoftMCL can still converge quickly within 20,000 steps. This is mainly determined by the contrastive samples introduced by the momentum queue. The soft samples can be stored and applied for training in other batches. The higher the quality of the contrastive samples the momentum queue provided, the better the representation can be learned by the SoftMCL.

\section{Conclusion}

This study proposed a soft momentum contrastive learning for sentiment-aware pre-training on both the word- and sentence-level to enhance the PLM's ability to learn affective information. The continuous affective ratings are adopted to guide the training of CL precisely. A momentum queue was introduced to expand the contrastive samples, allowing storing and involving more negatives to overcome the limitations of hardware platforms. Extensive experiments on four different sentiment-related tasks have proven the effectiveness of the proposed SoftMCL method.

Future works will attempt to incorporate the sentiment information into decoder-only architecture to achieve the sentiment expression ability for generative models.

\nocite{*}
\section*{Acknowledgements}\label{sec:acknowledgement}

This work was supported by the National Natural Science Foundation of China (NSFC) under Grant Nos. 61966038 and 62266051, the Ministry of Science and Technology (MOST), Taiwan, ROC, under Grant No. MOST110-2628-E-155-002. The authors would like to thank the anonymous reviewers for their constructive comments.

\section*{Bibliographical References}\label{sec:reference}

\bibliographystyle{lrec-coling2024-natbib}
\bibliography{lrec-coling2024-example}

\begin{appendices}
\section{Related Works}

The appendix briefly reviewed the related works of sentiment embeddings and sentiment-aware pre-training.

\vspace{0.5cm}

\subsection{Sentiment-enhanced Embeddings}

Using distributed representation to map words into latent space, the technique of word embeddings could make semantically related words appear closer to one another. These word embeddings can serve as the input for various neural network models to leverage the contextual information in large corpora, thus benefiting many NLP tasks \cite{Press2019, Wang2020a}. Unfortunately, the traditional semantic word embeddings are not sufficiently practical or feasible when directly used for sentiment analysis tasks. Several studies \cite{Agrawal2018} have reported that good is very similar to bad in vector space because these words are commonly used in similar contexts despite being opposite in sentiment. As a result, two sentences with similar contexts, respectively containing good and bad, will be assigned a similar sentence representation.

Previous studies have suggested using sentiment-aware pre-training \cite{Abdalla2019, Fu2018, Lan2016, Ren2016, Tang2016a} or sentiment refinement \cite{Utsumi2019, Yu2018b} to encode external knowledge into word representation to enrich the model with sentiment information.

\noindent \textbf{Sentiment-aware Pre-training.} The C\&W model \cite{Collobert2008, Collobert2011} is typically extended to train on a corpus with sentiment labels to learn sentiment embeddings from scratch. \citet{Tang2016a} proposed an end-to-end architecture with different loss functions to incorporate sentiment and semantic information of words. \citet{Ren2016} further integrated topic information into sentiment embeddings to learn a multi-prototype topic and sentiment-enriched word embeddings. \citet{Felbo2017} used a bi-directional LSTM model trained from scratch on almost one billion tweets to obtain context-aware sentiment embeddings.

\noindent \textbf{Sentiment Refinement.} An alternative method is to adjust pre-trained word embeddings with external word- or sentence-level knowledge using an embedding refinement model and then leverage the modified embeddings for sentiment analysis.

The refinement model could use affective lexicons to enrich word-level sentiment information to pre-trained word embeddings. \citet{Yu2018} and \citeyearpar{Yu2018b} extracted VA ratings from affective lexicons to pull similar sentiment words together in the latent space. \citet{Ye2018} used a joint learning approach to integrate sentiment into word embeddings by predicting both word- and sentence-level sentiment scores.

\vspace{0.5cm}

\subsection{Sentiment-aware Transformers}

For sentiment-related tasks, several studies proposed to train a transformer-based model by being supervised with sentiment labels as either a pre-training or a post-training task. For example, \citet{Xu2019b} proposed BERT-PT, which conducts sentiment-related post-training on the corpora to benefit aspect-level sentiment analysis. \citet{Tian2020} proposed SKEP to introduce three predictions with sentiment knowledge to learn a unified sentiment representation for various sentiment applications. \citet{Zhou2020b} proposed SENTIX to extract domain-invariant sentiment features from large-scale review datasets, which can then perform cross-domain sentiment classification without fine-tuning. \cite{Yin2020} proposed SentiBERT, which applied a bi-attention mechanism on top of the BERT representation to learn phrase-level compositional semantics. Ke et al. (2020) proposed SentiLARE, which adopted a label-aware masked language modeling to learn sentiment-aware language representation. \cite{Li2021c} proposed SCAPT to obtain implicit and explicit sentiment polarity by aligning the representation of implicit sentiment expressions to those with the same polarity.

\end{appendices}

\end{document}